\documentclass{article}
\usepackage{loping_temp,times}

\usepackage{color}
\usepackage[linkcolor=blue]{hyperref}
\usepackage{url}
\usepackage{graphicx}
\usepackage{dsfont}
\usepackage[ruled,linesnumbered,noend,noline]{algorithm2e}
\usepackage{ulem}
\usepackage{amsmath}
\usepackage{amsfonts}
\usepackage{marvosym}
\usepackage{bm}
\usepackage{booktabs}

\title{Realistic Surgical Simulation from Monocular Videos}
\author{
Kailing Wang\textsuperscript{*},
Chen Yang\textsuperscript{*},
Keyang Zhao,
Xiaokang Yang,
Wei Shen\textsuperscript{\Letter} 
\\
MoE Key Lab of Artificial Intelligence, AI Institute, Shanghai Jiao Tong University \\
\small{\textsuperscript{*} {Equal Contribution}}
}

\begin{document}

\newcommand{\name}{SurgiSim}

\maketitle

\begin{abstract}
This paper tackles the challenge of automatically performing realistic surgical simulations from readily available surgical videos. Recent efforts have successfully integrated physically grounded dynamics within 3D Gaussians to perform high-fidelity simulations in well-reconstructed simulation environments from static scenes.
However, they struggle with the geometric inconsistency in reconstructing simulation environments and unrealistic physical deformations in simulations of soft tissues when it comes to dynamic and complex surgical processes.
In this paper, we propose \name, a novel automatic simulation system to overcome these limitations.
To build a surgical simulation environment, we maintain a canonical 3D scene composed of 3D Gaussians coupled with a deformation field to represent a dynamic surgical scene. This process involves a multi-stage optimization with trajectory and anisotropic regularization, enhancing the geometry consistency of the canonical scene, which serves as the simulation environment.
To achieve realistic physical simulations in this environment, we implement a Visco-Elastic deformation model based on the Maxwell model, effectively restoring the complex deformations of tissues.
Additionally, we infer the physical parameters of tissues by minimizing the discrepancies between the input video and simulation results guided by estimated tissue motion, ensuring realistic simulation outcomes. 
Experiments on various surgical scenarios and interactions demonstrate \name's ability to perform realistic simulation of soft tissues among surgical procedures, showing its enormous potential for enhancing surgical training, planning, and robotic surgery systems. The project page is at \href{https://namaenashibot.github.io/SurgiSim/}{https://namaenashibot.github.io/SurgiSim/}.
\end{abstract}
\section{Introduction}
Realistic surgical simulation systems are pivotal in enhancing clinical training, offering substantial benefits for training surgeons, and advancing the development of robotic surgery systems~\cite{intro1,intro2}.
Currently, mainstream surgical simulation systems such as LaparoS$~^{\text{\tiny TM}}$ from VIRTAMED are commercial products that primarily utilize mesh-based technology.

\begin{figure}[htb]
\begin{center}
\includegraphics[width=0.45\textwidth]{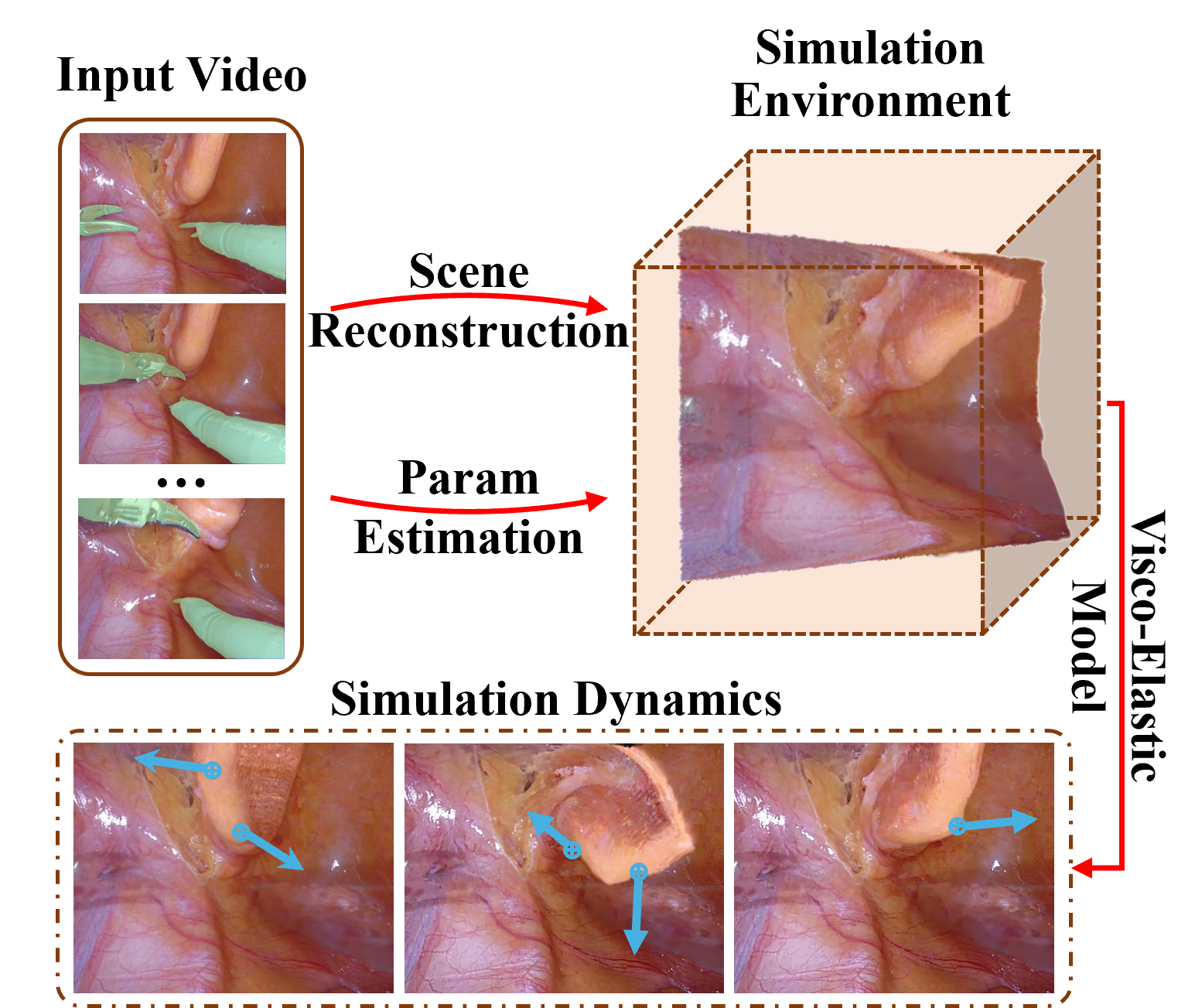}
\end{center}
\caption{An overview of \name. From an input surgical video, we first reconstruct a 3D scene as the simulation environment, then estimate its physical parameters based on a Visco-Elastic deformation model, and finally perform realistic simulation dynamics in it. The green parts are surgical tool masks and the blue arrows indicate the directions of applied forces in simulations.
}
\label{fig:pipeline}
\end{figure}

While these systems provide highly realistic simulations, their performance and reliability heavily depend on labor-intensive mesh building and physical parameter tuning. The dependence on the skills of modelers and animators limits the variety of possible simulation scenarios. 
In this paper, we aim to develop a high-quality and flexible simulation system capable of automatically converting real surgical videos into surgical simulation environments as well as performing high-fidelity and realistic simulations.

Building high-quality surgical simulation systems from readily available surgical videos poses significant challenges, including the reconstruction of simulation environments from monocular videos with complex rigid and non-rigid deformations and accurately modeling the soft tissues for realistic surgical simulation. 
Fortunately, recent advancements in scene reconstruction and physically grounded dynamics offer plausible solutions. Methods based on Neural radiance fields (NeRF)~\cite{endonerf, lerplane,forplane} and 3D Gaussian Splatting (3DGS)~\cite{endogslam,deform3dgs,endogaussian} have demonstrated impressive capabilities in reconstructing surgical scenarios. 
Besides, PhysGaussian~\cite{physgaussian} combines continuum mechanics with 3DGS and employs the Material Point Method (MPM) to simulate realistic motions in reconstructed models under simple point load. 
However, despite these technological strides, two significant challenges still obstruct the realization of the target system.

Firstly, current methods cannot \textbf{reconstruct geometrically consistent surgical simulation environments}.
Recent advances in dynamic tissue reconstruction have shown promising capabilities by modeling tissues using canonical scenes with deformation fields~\cite{surgicalgaussian,endogaussian,endonerf,lerplane,forplane}. However, these methods typically treat dynamic tissue reconstruction as a task of synthesizing high-fidelity images at specific timestamps.
Under this definition, the deformation field is optimized to produce visually plausible results for each individual frame rather than maintaining consistent geometry across time. Consequently, the geometry mapping between the canonical scene and different timestamps becomes physically implausible. In surgical simulations, these geometric inconsistencies lead to significant artifacts such as fragmentation and messy deformations.
Thus, it is crucial to develop a method capable of efficiently transforming surgical videos into high-quality, geometrically consistent, and simulation-ready environments.

Secondly, current methods fail to \textbf{simulate complex tissue deformations}.
Current representative simulation methods, such as PhysGaussian~\cite{physgaussian}, primarily employ a Plastic-Elastic model with uniform simulation parameters assigned manually across the entire subject. This design falls short in capturing the complicated deformation behaviors inherent 
in real surgical scenarios. 
Moreover, the manual assignment of physical parameters to the simulation subject compromises the realism of the simulation.
Recent research~\cite{physdreamer,physics3d} propose to leverage diffusion priors, either in the form of generated videos or through score distillation sampling (SDS)~\cite{dreamfusion} to estimate physical parameters and avoid the need for manually assigned parameters. While these diffusion-based methods are effective in simple scenarios such as handling natural fluctuation, they fail in more complex surgical situations that involve external forces from operations like pulling or cutting. Moreover, current medical video generation models~\cite{endora,bora} are still in the early stages of development~\cite{surgen}, and the quality of their output is insufficient to provide guidance for simulations. 

To address the aforementioned challenges, we propose~\name, an automated system for realistic, high-quality surgical simulation leveraging readily available surgical videos. \name~incorporates a novel module that employs 3DGS~\cite{3dgs}, an explicit 3D representation capable of real-time rendering, to construct a canonical scene from surgical video footage. This module utilizes multi-stage optimization with trajectory and anisotropic regularization to enhance geometry consistency, complemented by a surface thickening method to enrich reconstructed tissue content for improved simulation fidelity. 
To better model tissue deformations across various surgical scenarios, we implement a Visco-Elastic deformation model based on the Maxwell model~\cite{maxwell}. We further employ a physics-guided parameter estimation method to acquire physical parameters of tissues, leveraging observed physical effects in the input video. Specifically, we utilize point tracking~\cite{dot} and depth estimation~\cite{dav2} to derive the tissue motions induced by external forces, with physical parameters optimized through differentiable simulation and rasterization. This method ensures high-quality simulation with realistic physical effects, closely mimicking the complexities of real surgical scenarios. 
Extensive experiments across diverse surgical procedures demonstrate \name's capability to realistically simulate soft tissue interactions, highlighting its potential for enhancing surgical training, planning, and robotic surgery systems.

We summarize our contributions as follows.
\begin{enumerate}
    \item An automatic system that takes only a monocular surgical video as input to a) build a simulation-ready surgical environment with physics parameters estimated based on the input video and b) perform realistic physics-based simulation in it.
    \item A novel multi-stage optimization strategy for 3DGS that ensures geometric consistency across temporal deformations, enabling reliable simulation environments.
    \item A Visco-Elastic deformation model, coupled with automated physical parameter estimation from video observations using point tracking and depth estimation, facilitates high-fidelity and realistic surgical simulations.
\end{enumerate}

\section{Related Work}

\paragraph{Dynamic Scene Representations.}
The rapidly advancing field of Neural Radiance Filed (NeRF \cite{nerf}) and 3DGS~\cite{3dgs} has garnered significant interest for dynamic scene reconstruction. 
Early works based on NeRF~\cite{4dnerf1,4dnerf2,4dnerf3,4dnerf4} model scenes with implicit representations, which are difficult to interact with and therefore unsuitable for simulation. Later works based on 3DGS~\cite{4dgs1,4dgs2,4dgs3,4dgs4,4dgs5} use explicit representation, but require multi-view images or videos with severe movement of cameras, which is hard to obtain in the field of surgery. 
Recently, some other research \cite{endonerf,lerplane,forplane,deform3dgs,endogaussian,surgicalgaussian,endo4dgs,endogs,freedygs} has focused on the reconstruction of dynamic surgical scenes. However, these studies either still use implicit representation, or primarily aim to reconstruct the dynamic scenes at specific timestamps, instead of the geometric continuity and consistency of the reconstructed canonical model.

\paragraph{Material Point Method.}
Material Point Method is a hybrid Lagrangian/Eulerian discretization scheme for solid mechanics~\cite{mpm0}. The MPM system's inherent ability to handle topology changes and frictional interactions makes it well-suited for simulating a wide range of materials~\cite{physgaussian}.
including elastic objects, sand, cloth, hair, snow and lava~\cite{mpm25,mpm1,mpm2,mpm3,mpm4,mpm5,mpm6}. In addition, modern implementations of MPM utilize the parallel ability of GPUs to achieve advanced efficiency\cite{mpmgpu}. 
Recently, \citet{physgaussian} uses GPU-based MPM to efficiently incorporate dynamics into different scenarios using a unified particle representation within the Gaussian Splatting framework. However, their methods rely on manually assigned physical parameters, and assume the parameters to be the same over large regions. Follow-up works by~\citet{physdreamer,physics3d} further utilize the differentiable ability of the simulation process to estimate the parameters in MPM guided by Diffusion Models~\cite{svd}. These methods extract the physical prior from generative models, which is unreliable in the surgery field.

\paragraph{Surgical Simulation Systems.}
Medical simulation systems are based on the deformation models used to handle tissue motion under the interaction of external forces~\cite{simulation0}. Early systems apply mainly heuristic approaches like deformable spines, spring-mass models or linked volumes~\cite{simulation1,simulation2,simulation3,simulation4}. These systems are limited by the computing hardware and only perform simple simulations. 
With the development of computer graphics technology, later methods and products either base the simulation on manually built mesh models and predefined animation or replay real videos captured for VR simulators~\cite{simulationanimation,simulationvr}. They require expensive human labor or advanced hardware or VR video capture but provide limited interaction. Recently, \citet{simendogs} proposed SimEndoGS, a data-driven system based on 3DGS and MPM. However, they use an elastic model to capture tissue motion and only support minor interactions like tiny force impulses.
\section{Method}
We propose~\name~(as illustrated in Fig. \ref{fig:pipeline}), an automatic system for surgical simulation, which constructs realistic surgical scenes as well as performs realistic simulations. The input of the system is a monocular surgical video with obvious tissue motion caused naturally or by surgical instruments. The system will reconstruct a simulation environment with physical parameters inferred from the video, and then perform high-fidelity simulations.
The following sections detail the methodology: 
Sec.~\ref{prelim} details foundational techniques. 
Sec.~\ref{staticrecon} describes the reconstruction of a high-quality canonical simulation environment from a surgical video using a multi-stage optimization process with trajectory and anisotropic regularization.
Sec.~\ref{realsim} demonstrates how we model the complex deformations of tissues, thereby facilitating high-quality simulations.

\subsection{Preliminaries}
\label{prelim}

\paragraph{3D Gaussian Splatting.}
We use 3D Gaussian Splatting~\cite{3dgs}, a novel differentiable rendering method which represents scenes with collections of anisotropic 3D Gaussian Kernels $\mathcal G = \{G_i: \mu_i, o_i, \Sigma_i, C_i\}_{i=1}^N$, where $\mu_i, o_i, \Sigma_i, C_i$ represents the position, opacity, covariance matrix, and spherical harmonic (SH) coefficients of the $i_{th}$ Gaussian from all $N$ Gaussian kernels. The covariance matrix $\Sigma$ is further decomposed into rotation matrix $R$ and scaling matrix $S$. To render an image through the differentiable rasterization of 3DGS, 3D Gaussian kernels will be projected onto the image plane, and the RGB color is computed as
\begin{equation}
    \mathbf{\hat{C}} = \sum_{i \in \{N\}} \alpha_i \operatorname{SH}(d_i, C_i) \prod_{j=1}^{i-1} (1-\alpha_j), 
    \label{rendereq}
\end{equation}
where SH denotes the computation of color values based on the given view and SH coefficients, $d_i$ is the view direction from the camera to $G_i$, and $\alpha_i$ represents the effective opacity ordered by z-depth, calculated by multiplying the 2D Gaussian weight with each point's inherent opacity $o_i$.
To integrate 3DGS into the simulation pipeline, we view Gaussian kernels as particles carrying properties.

\paragraph{Continuum Mechanics.}
Continuum mechanics models motion with a transformation map $\mathbf x = \phi(\mathbf X, t)$, where $\mathbf x$ represents a material point in the world space $\Omega^t$ at time $t$, deformed from point $\mathbf X$ in the undeformed material space $\Omega^0$. The deformation gradient, $\mathbf F = \frac{\partial \phi}{\partial \mathbf X}$, describes local motion and strain~\cite{mpmbook}.
In continuum mechanics, the two primary constraints are mass conservation and momentum conservation, given by 
\begin{equation}
    \int_{\Omega_\epsilon^t} \rho(\mathbf x, t) = \int_{\Omega_\epsilon^0} \rho(\mathbf X, 0),\text{~}\rho(\mathbf x, t)\overset{.}{\mathbf v}(\mathbf x, t) = \nabla \cdot \bm \sigma(\mathbf x, t) + \mathbf f^{\text{ext}},
\end{equation}
where $\Omega_\epsilon^t \in \Omega^t$ is an infinitesimal region, $\rho$ and $\mathbf v$ denote the density and velocity filed respectively, and $\mathbf f^{\text{ext}}$ is the external force. $\bm \sigma$ is the Cauchy stress tensor, usually related to a given energy function $\Psi$. %  Details on $\bm \sigma$ will be introduced in Sec. \ref{realsim}. 

\paragraph{Marerial Point Method.}
Material Point Method (MPM) discretizes the continuum into a collection of Lagrangian particles. The mass conservation of each particle ensures the overall mass conservation. Following~\citet{mpmsnow}, we use particle-to-grid (P2G) and grid-to-particle (G2P) to transfer properties between these particles and an Eulerian grid. The momentum conservation is ensured on the grid where the calculation is simpler and more natural. In each simulation step, the mass and momentum are first transferred from particles onto the grid. The stress tensor is used to update the grid velocities, and the velocities will be transferred back to update particle states. The velocities on the grid are updated as 
\begin{equation}
    \mathbf v_i^{n+1} = \mathbf v_i^n - \frac{\Delta t}{m_i} \cdot \operatorname{P2G}(\{\bm\sigma\}),\text{~}\mathbf v_j^{n+1} = \operatorname{G2P}(\{\mathbf v_i^{n+1}\})
\end{equation}
where $\operatorname{P2G}$ is the transfer of stress from particles to the grid and $\operatorname{G2P}$ is the reverse transfer. $n$ denotes the $n$-th simulation step, each lasts for $\Delta t$, and $\mathbf v_i,\text{~}\mathbf v_j$ denote velocity vectors on the $i$-th grid node and $j$-th particle respectively. Gravity is ignored in our implementation.
Please refer to the supplementary materials for further details on MPM.

\subsection{Simulation environment setup}
\label{staticrecon}
In this section, we first describe the process of constructing a static simulation environment (canonical 3D scene) from a dynamic surgical video. Subsequently, we detail the regularization techniques designed to enhance geometric consistency, along with the surface thickening methods used to complete the canonical scene.

\paragraph{Data Preparation.}
Given a monocular RGB surgical video $\{\mathbf C_o^t\}^T_{t=1}$ with $T$ frames, we first use SAM~\cite{samv1} segment frames to generate tissue masks $\{\mathbf M^t\}^T_{t=1}$. Then we employ a video inpainting method~\cite{e2fgvi} to inpaint the RGB frames on the areas covered by surgical instruments with $\{\mathbf M^t\}^T_{t=1}$. 
These inpainted images $\{\mathbf C^t\}^T_{t=1}$ are input to Depth Anything v2 model~\cite{dav2} to estimate depth $\{\mathbf{\hat D}^t\}^T_{t=1}$.

\paragraph{Canonical Scene Reconstruction.}
The frames and their corresponding depths from data preparation are used to build a canonical 3D scene composed of 3D Gaussians, which later serves as a simulation environment.
Inspired by~\citet{4dgs1,endogaussian}, we use a deformation field $\mathbf D$ to model the 4D deformation of a canonical 3D Gaussian model $\mathcal G^0$. 
To achieve this, we start by initializing a coarse 3D Gaussian model using RGBD projection using the inpainted first frame and the corresponding estimated depth. Pixels in other frames will be projected into the coarse 3D Gaussian model if they belong to the mask area of all previous frames. 
The deformation field $\mathbf D$ is composed of a set of multi-resolution feature planes $\{\mathbf V_{ij}\} \subset \mathcal R^{h\times lN_i \times lN_j}$ and an MLP $\bm \theta$, where $h$ is the hidden feature size, $l$ is the upsampling scale and $N$ is the resolution parameter. To query the deformation of a Gaussian kernel $G_k: \mu_k, \alpha_k, \Sigma_k, C_k$, we first calculate the voxel feature:

\begin{equation}
    f_v = \bigcup_l \operatorname{lerp}(\mathbf V_{ij}, \mu_k),\text{~}ij \in \{xy,xz,xt,yz,yt,zt\}.
\end{equation}

Here $\operatorname{lerp}$ denotes 4-nearest bilinear interpolation. Then the feature is decoded by the MLP, 
\begin{equation}
    \Delta \mu, \Delta o, \Delta R, \Delta S = \bm \theta (f_v),
\end{equation}
and the attributes of the deformed Gaussian can be computed as 
\begin{equation}
    G_k' = (\mu_k+\Delta\mu, o_k'+\Delta o, \Sigma(R_k+\Delta R, S_k+\Delta S), C_k).
\end{equation}

Different from previous methods~\cite{4dgs1,endogaussian}, we allow the opacity to change during deformation. The opacity of tissues would change significantly when we pull it hard, unlike common objects.
To render the deformed Gaussian model at a certain timestamp for optimization, we use the deformed attributes calculated above for rasterization in Eq.~\ref{rendereq}.

\paragraph{Multi-Stage Optimization.}
Our key goal is to create a surgical scene that allows for high-quality surgical simulation rather than generating high-fidelity images at specific timestamps, which all previous tissue reconstruction methods \cite{lerplane,forplane,endogaussian,surgicalgaussian} aimed for.
Thus, our design focuses on maintaining the geometric consistency of the canonical Gaussian model $\mathcal G^0$ during the deformation process.
Specifically, we design an explicit trajectory regularization strategy to prevent positional interleaving during deformation, which otherwise leads to faulty optimization of the canonical model.
This regularization requires that the trajectories of Gaussian points within any small region tend to be parallel within a short period of time.
For a Gaussian model $\mathcal G$, we first find the $k$-nearest neighbor point set $N_i$ for each Gaussian $G_i$. For a certain timestamp $t$, the regularization is given by
\begin{align}
    \mathcal L_{traj} &= \sum_{G_j, G_k \in N_i} \frac{\Delta \bm\mu_j^t \cdot \Delta \bm\mu_k^t}{\lVert \Delta \bm\mu_j^t \rVert \cdot \lVert \Delta \bm\mu_k^t \rVert} \cdot \lVert \Delta \bm\mu_j^t \rVert \cdot \lVert \Delta \bm\mu_k^t \rVert \notag \\
    &= \sum_{G_j, G_k \in N_i} \Delta \bm\mu_j^t \cdot \Delta \bm\mu_k^t.
    \label{trajloss}
\end{align}
where $\Delta \bm \mu_i^t$ means the deformation vector from the previous timestamp, i.e., $\mu^t_i - \mu^{t-1}_i$. 
This regularization consists of two main parts. The first part requires the deformation direction to be parallel to avoid tangential misalignment in the direction of motion. The second part requires the movement length to be small to avoid radial misalignment. 

The multi-stage trajectory optimization contains three stages:
1) Optimizing only the canonical model $\mathcal G^0$ and the deformation field $\mathbf D$ (the feature planes and the MLP) to estimate the coarse trajectory.
2) Adding the trajectory regularization to refine the trajectory of all Gaussian kernels.
3) Freezing the deformation module and only optimizing the attributes of the canonical model $\mathcal G^0$.

Additionally, an anisotropic regularization proposed to prevent excessively large or extremely anisotropic Gaussian kernels is employed across all the stages mentioned above. It is defined as:
\begin{align}
    \mathcal L_{geo} &= \sum_{i \in \{N\}} \left( \operatorname{ReLU}\left(\max(S_i) - r_m\right) \right. \notag \\
    &\quad + \left. \operatorname{ReLU}\left( \max(S_i)/\min(S_i) - r_{ani}\right) \right)
\end{align}
where $r_m=1$ is the max scaling limit, and $r_{ani}=3$ is the anisotropic factor.

\paragraph{Surface Thickening.} 
Due to the limited camera view, the reconstructed canonical model $\mathcal G^0$ is a single surface with invalid thickness and volume for simulation. To address this, we apply a surface thickening method, pushing the Gaussian kernels in $\mathcal G^0$ along the z-axis with a certain probability. The thickening algorithm is described in Algo. \ref{alg:densefill}.

\begin{algorithm}[h]
\small
\caption{Surface Thickening Algorithm}
\label{alg:densefill}
\SetKwInOut{Input}{Input}
\SetKwInOut{Output}{Output}
\Input{The canonical Gaussian model $\mathcal G^0$}
\Output{A thickened model for simulation $\mathcal G_\text{d}^0$}
Initialize $\mathcal G_\text{d}^0 \leftarrow \mathcal G^0$; \\
$\{x_\text{m}, x_\text{M}, y_\text{m}, y_\text{M}, z_\text{m}, z_\text{M}\} \leftarrow$ bounding box of $\mathcal G^0$; \\
\For{\textnormal{Layer $l$ in range(1, 1000})}
{   
    \For{\textnormal{Each Gaussian kernel $G^0_i$ in $\mathcal G^0$}}
    {
        $G^l_i \leftarrow G^0_i$; \\
        The position of $G^l_i$: $\mu^l_i \leftarrow \mu^0_i \cdot (\operatorname{rand}(3)+l)/1000$; \\
        \If{\textnormal{$G^l_i$ in $\{x_\text{m}, x_\text{M}, y_\text{m}, y_\text{M}, z_\text{m}, (1+0.25)z_\text{M}\}$}}
        {$\mathcal G_\text{d}^0 \leftarrow \mathcal G_\text{d}^0 \cup \{G^l_i\}$ ;}
    }
}
return $\mathcal G_\text{d}^0$;
\end{algorithm}

\subsection{Physics-based Realistic Simulation}
\label{realsim}
In this section, we first introduce how we model the complex physical effects of tissues with the Visco-Elastic model. Then we demonstrate how we use this model and input video to infer the physical parameters.
For better understanding, we employ conventional notations from physics. This may lead to some repetition of the previously defined notations. These corrupted definitions are only used in this section, and their meaning will be re-defined. 

\begin{figure*}[ht]
\begin{center}
\includegraphics[width=0.85\textwidth]{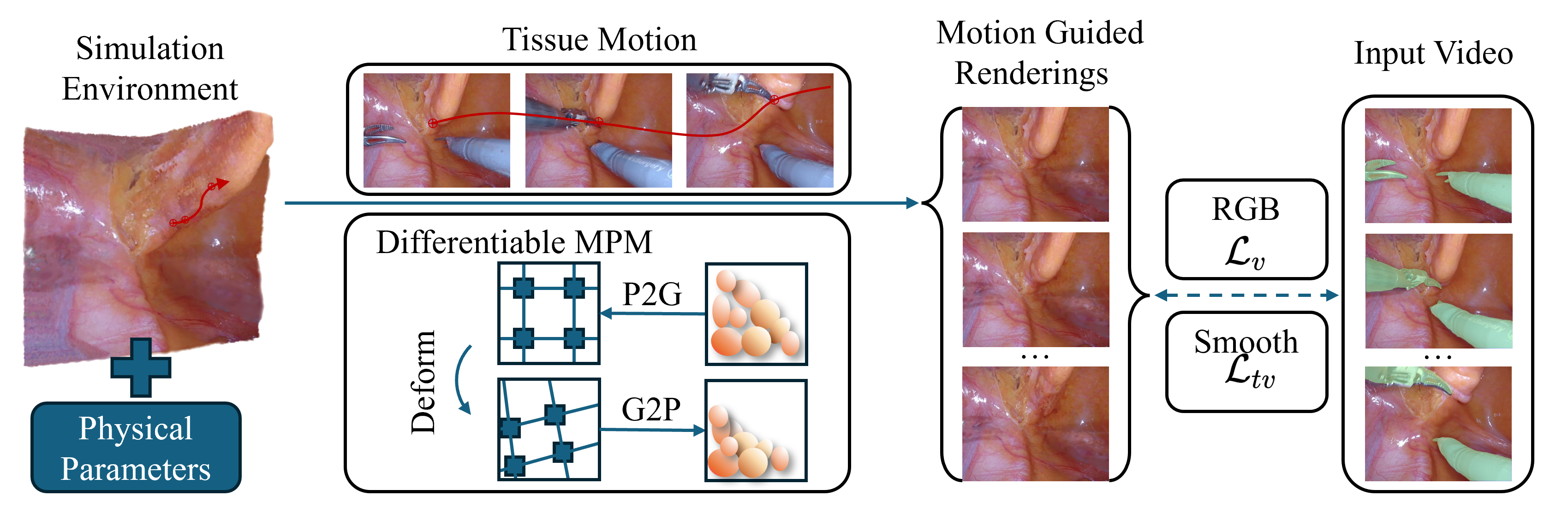}
\end{center}
\caption{Illustration of our physical parameter estimation. \name~automatically infers physical parameters by minimizing the discrepancy between rendered simulation results and the input video through differentiable MPM and rasterization.}
\label{fig:physes}
\end{figure*}

\paragraph{Modeling Tissues with Visco-Elasticity Deformation.}
Previous methods~\cite{physgaussian,physdreamer} primarily employ elastic models for simulation, which restricts their performance to jelly-like effects that oscillate back and forth. However, in surgical settings, due to the complex composition of tissues, their deformations are usually very complicated in physical behaviors, which significantly differ from these jelly-like ones. To better model these specific effects, we involve a Visco-Elastic model~\cite{nrslam}, which consists of an elastic part and a viscous part to model the complex deformations.

For the elastic part, we use fixed corotated elasticity. The energy function is defined as:
\begin{equation}
    \Psi(\mathbf F_{\mathbf E}) = \Psi(\bm \Sigma_{\mathbf E}) = \mu_{\mathbf E} \sum_i (\sigma_{\mathbf E,i}-1)^2 + \frac{\lambda_{\mathbf E}}{2}(\operatorname{det}\mathbf F_{\mathbf E}-1)^2,
\end{equation}
where the elastic deformation gradient $\mathbf F_{\mathbf E}$ is decomposed into $\mathbf U_{\mathbf E} \bm \Sigma_{\mathbf E} \mathbf V_{\mathbf E}^\top$ through SVD, $\sigma_{\mathbf E, i}$ are the singular values of $\mathbf F_{\mathbf E}$. $\mu_{\mathbf E} = \frac{E}{2(1+\nu)}$ and $\lambda_{\mathbf E} = \frac{E\nu}{(1+\nu)(1-2\nu)}$ are physical parameters, namely Shear modulus and Lam\'{e} modulus, computed from the Young's modulus $E$ and Poisson's ratio $\nu$. The Cauchy stress tensor for the elastic part is $\bm \sigma_{\mathbf E} = \frac{1}{\operatorname{det} \mathbf F} \frac{\partial \Psi}{\partial \mathbf F_{\mathbf E}}(\mathbf F_{\mathbf E})\mathbf F^\top_{\mathbf E}$. The update rule of $\mathbf F_{\mathbf E}$ is 
\begin{equation}
    \mathbf F_{\mathbf E, j}^{n+1} = \mathbf F_{\mathbf E, j}^n(\mathbf I + \Delta t \cdot \nabla \mathbf v_j),
\end{equation}
and $j$ denotes the $j$-th particle.

As for the viscous part, we get inspiration from~\citet{simo},~\citet{maxwell} and propose a simple model regarding viscous energy dissipation to integrate it into the MPM. The viscous energy dissipation is described through a dissipation potential, given by
\begin{equation}
    \Psi(\frac{\partial \mathbf F_{\mathbf v}}{\partial t}) = \frac{1}{2} \eta_{\mathbf v} \operatorname{tr}\left(\left(\frac{\partial \mathbf F_{\mathbf v}}{\partial t}\right)^\top \frac{\partial \mathbf F_{\mathbf v}}{\partial t}\right),
\end{equation}
where $\eta_{\mathbf v}$ is the viscosity coefficient. The gradient of the viscous deformation gradient $\mathbf F_{\mathbf v}$ can be updated by $\frac{\partial \mathbf F_{\mathbf v}}{\partial t} = \gamma\cdot\mathbf D$, and $\mathbf D = \frac{1}{2}(\nabla \mathbf v_j + \nabla \mathbf v_j^\top)$ is the symmetric part of the grid velocity. The viscous Cauchy tensor is $\bm \sigma_{\mathbf v} = \operatorname{det}(\mathbf F_{\mathbf v}) \cdot 2 \eta \mathbf D$, and $\mathbf F_{\mathbf v}$ can be updated by
\begin{equation}
    \mathbf F_{\mathbf v, j}^{n+1} = \mathbf F_{\mathbf v, j}^{n}(\mathbf I + \gamma_{\mathbf v}\Delta t\cdot\mathbf D).
\end{equation}

The overall stress tensor $\bm \sigma = \bm \sigma_{\mathbf E}+\bm \sigma_{\mathbf v}$ is used to update the grid velocity.

\paragraph{Tissue Motion Estimation.}
To estimate physical parameters, we involve extracting the physical priors of tissues from surgical videos. This process comprises two main steps: 1) recovering the tissue motion induced by external forces from instruments as captured in the input video, and 2) refining physical parameters by minimizing the discrepancies between the simulated dynamics guided by the tissue motion and input video.

One straightforward way to recover motion in MPM simulation is to estimate the forces involved. However, due to uncertainties in system dynamics, accurately estimating forces is a challenging task \cite{forceestm}. Instead, we turn to estimating the 3D trajectory of tissues. We start by selecting pixels on the tissues near the contact point with the surgical instruments and employ a 2D dense optical tracking method \cite{dot} to capture the 2D trajectories of tissue movements directly influenced by the external forces. We then augment these 2D trajectories with estimated depth values, $\{\mathbf{\hat D}^t\}^T_{t=1}$, to derive 3D trajectories, $\{\mathbf p^t\}^T_{t=1}$.

To accurately recover motion in the video, we update the velocity during MPM at time $t$. Specifically, we adjust the grid velocity in a small region $\mathcal B^0 \in \Omega^0$ around the starting point of the 3D trajectory $\mathbf p^0$. The velocity is updated according to the following formula:
\begin{equation}
    \mathbf v_{\mathcal B^0}^t = \frac{p^{n_t+1}-p^{n_t}}{\Delta T},
\end{equation}
where $\Delta T$ is the video frame duration and $n_t$ is the frame index that $n_t\Delta T \leq t < (n_t+1) \Delta T$. 

\paragraph{Physical Parameter Estimation.} 
We then use the input video to estimate the physical parameters of tissues.
This estimation is achieved by minimizing the discrepancies between frames from the input video and the simulation results driven by estimated motion at the time of the frames, as shown in Fig.~\ref{fig:physes}.

Firstly, we run $\frac{t}{\Delta t}$ simulation steps to get the simulated model $\mathcal G_d^t$ deformed from the dense model $\mathcal G_d^0$, and then rasterize the model as $\hat{\mathbf C^t_{s}}$ with Eq.~\ref{rendereq}. 
In this way, we align the simulation results with the input video.
We then optimize the parameters using
\begin{equation}
    \mathcal L_{v} = \lVert \mathbf C_o^t - \mathbf C_s^t \rVert_1 \cdot \mathbf M^t.
\end{equation}

Due to the influence of previous simulation steps on subsequent ones, we implement training in a rolling manner. Specifically, we begin by optimizing with the first $k$ video frames. For each new round, we start from the first guide frame and add additional $k$ frames until the guide length surpasses the video length.
A parameter smoothing is conducted every $k$ frame. For each neighbor group $N_i$ for each Gaussian $G_{d, i}$, we apply a total variation loss:
\begin{equation}
    \mathcal L_{tv} = \underset{G_{d, j}\in N_i}{\operatorname{MSE}} (\xi_j - \xi_i),
\end{equation}
where $\xi$ is one of the physical parameters, namely $\mu_{\mathbf E}$, $\eta_{\mathbf v}$ and $\gamma_{\mathbf v}$. 
We set $\nu_{\mathbf E}$ to a constant value of $0.45$, because we have found that variations within the valid range have little impact on the results.

\section{Experiments}

\begin{figure*}[t]
\begin{center}
\includegraphics[width=0.9\textwidth]{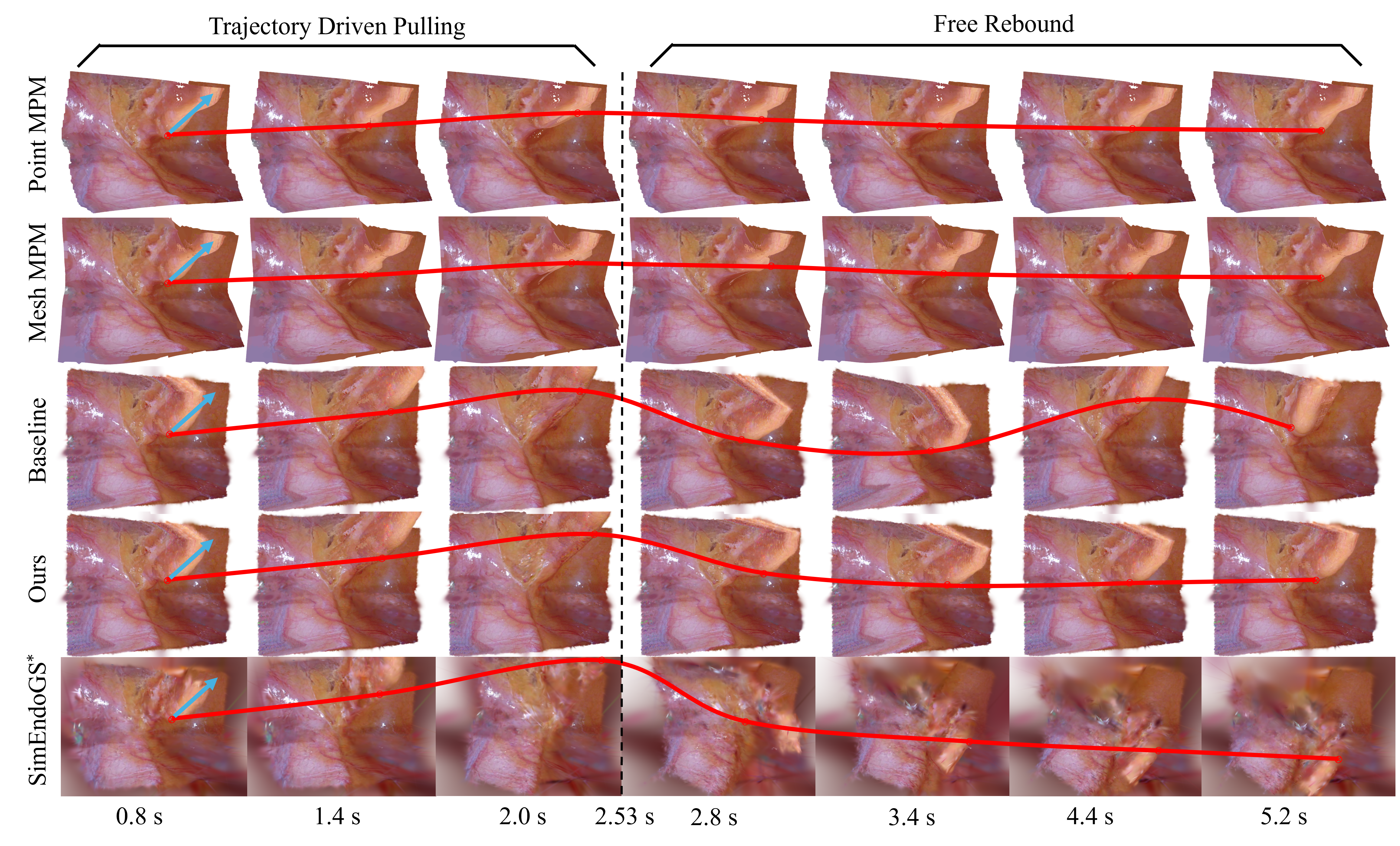}
\end{center}
\caption{Visualization of simulations. We show the trajectory direction with a blue arrow and the motion of the tissues with a red line. The external force caused by the driving trajectory ends at 2.52s (63 frames), after which the tissue rebounds freely. \name~consistently produces the most realistic simulation dynamics. Please refer to the supplementary material for more simulation results.}
\label{fig:simres}
\end{figure*}

\subsection{Implementation Details.}

\paragraph{Dataset.} We evaluate our method on the EndoNeRF dataset~\cite{endonerf}, which comprises several surgical video clips totaling 807 frames. Each clip, captured by stereo cameras from a single viewpoint, spans 4-8 seconds and shows typical soft tissue scenarios encountered in robotic surgery, including complex non-rigid deformations. We utilize 5 of these clips to establish our simulation environments, excluding one clip that involves significant tissue cutting. Though EndoNeRF is a binocular dataset, we use only the left image and no binocular depth. Please refer to supplement material for demos on more clips beyond EndoNeRF.

\paragraph{Environment and Simulation Setup.} Our environment setup employs a multi-stage optimization process consisting of 5000 iterations. The initial stage, comprising the first 50\% of iterations, focuses on optimizing the canonical model $\mathcal G^0$ and the deformation field $\mathbf D$. This is followed by a trajectory regularization stage from 50\% to 70\% of the total iterations. A refinement period then occurs from 70\% to 90\%, allowing for further optimization of the deformation field after its correction by trajectory optimization. In the final stage, occupying the last 10\% of iterations, we freeze the deformation field and fine-tune the canonical model. For the simulation, we adapt the MPM framework from \citet{physgaussian,physgaussianmpm}. The thickened model $\mathcal G_\text{d}^0$ is normalized into a 2-unit cube, overlaid with an Eulerian grid of resolution $50\times 50\times 50$. 
We established a conversion rate where 10,000 simulation steps correspond to one second of video at 25 fps. For each operation in the simulation, we conduct 80k simulation steps to form an 8-second video. To perform a simulation operation, physical parameters estimated from previous simulations are loaded onto the simulation scene for initialization.

\paragraph{User Study.} To evaluate the fidelity of our simulations, we conducted a user study involving 68 participants including both surgeons and laypersons. For More details, please refer to the supplementary material.

\subsection{Comparisons on Surgical Simulations}
We first provide qualitative comparisons on surgical simulation in Fig. \ref{fig:simres}.
We primarily compare with three simulation baselines:
1) A baseline using our simulation environment and native PhysGaussian \cite{physgaussian} as simulation backend, without our Visco-Elastic model and estimated parameters;
2) A pointcloud-based MPM approach, \textit{Pcd} for short;
3) A triangle mesh simulation with the same MPM, \textit{Mesh} for short. 
The MPM is built on the string-mass model from Taichi~\cite{taichi} framework.
All methods share the same manipulation during simulation.
In real world, due to the viscous properties of tissues \cite{nrslam}, under this manipulation, the tissue will quickly return to a stationary state rather than oscillating back and forth. The simulation results demonstrate that \name~achieves superior simulation results compared to baseline in both visual quality and realism. Our approach generates more realistic tissue deformation responses and interaction dynamics, closely mimicking the behavior observed in actual surgical scenarios. 
In contrast, the baseline often produces unrealistic elastic behavior that deviates from true tissue properties. Pcd and Mesh suffer from fragmentation, artifacts and unnatural motion during simulation. We also provide results of SimEndoGS$^\ast$, our own implementation of a SOTA method SimEndoGS \cite{simendogs}, which is not yet open-source. However, it does not perform proper simulation scene preparation as we do in Sec. \ref{staticrecon}, which causes it to fail under large movement in all cases for evaluation, and thus we left this method for further quantitative comparison.

\begin{figure}[tbt]
\begin{center}
\includegraphics[width=0.475\textwidth]{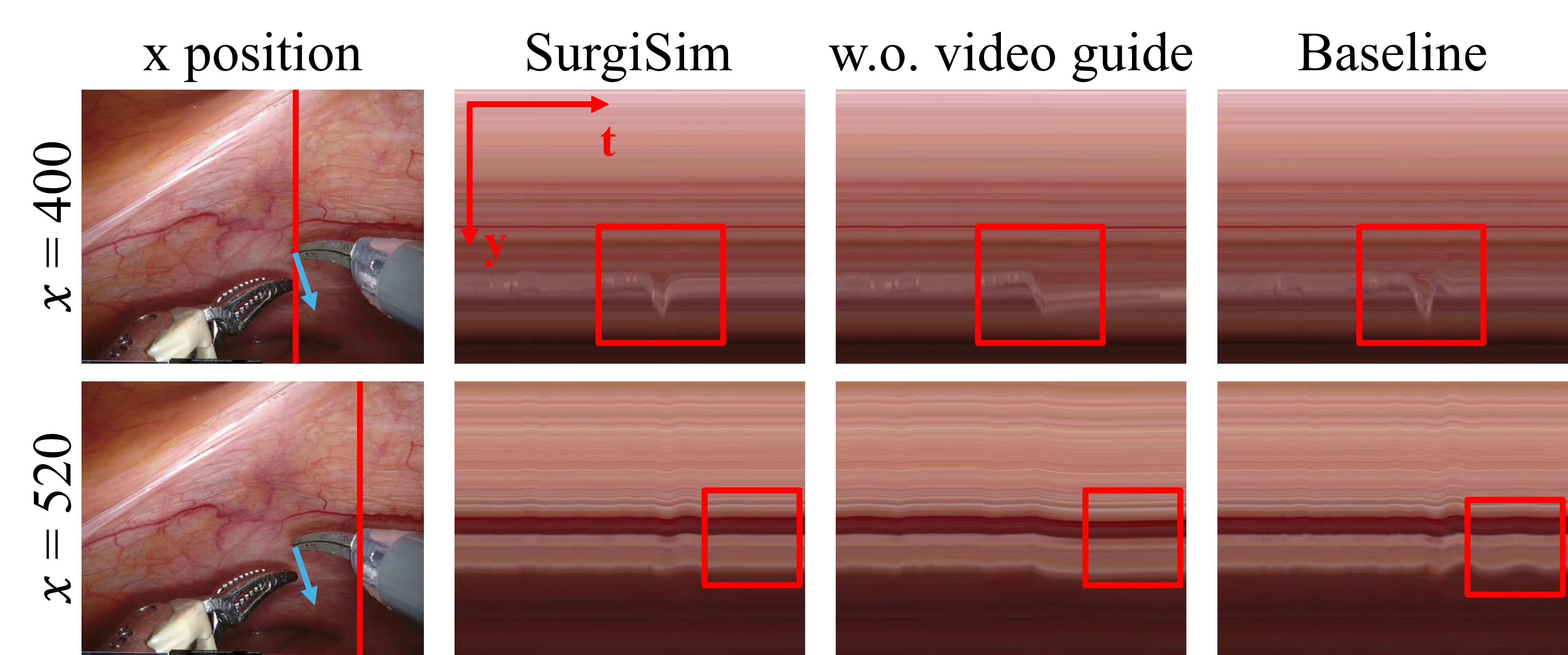}
\end{center}
\caption{Y-T slices of simulation dynamics. The slices at $x = 400$ capture the motion of tissues being lifted and then released. The slices at $x = 520$ capture the oscillations after the rebound.}
\label{fig:yt}
\end{figure}

\begin{figure*}[ht]
\begin{center}
\includegraphics[width=0.8\textwidth]{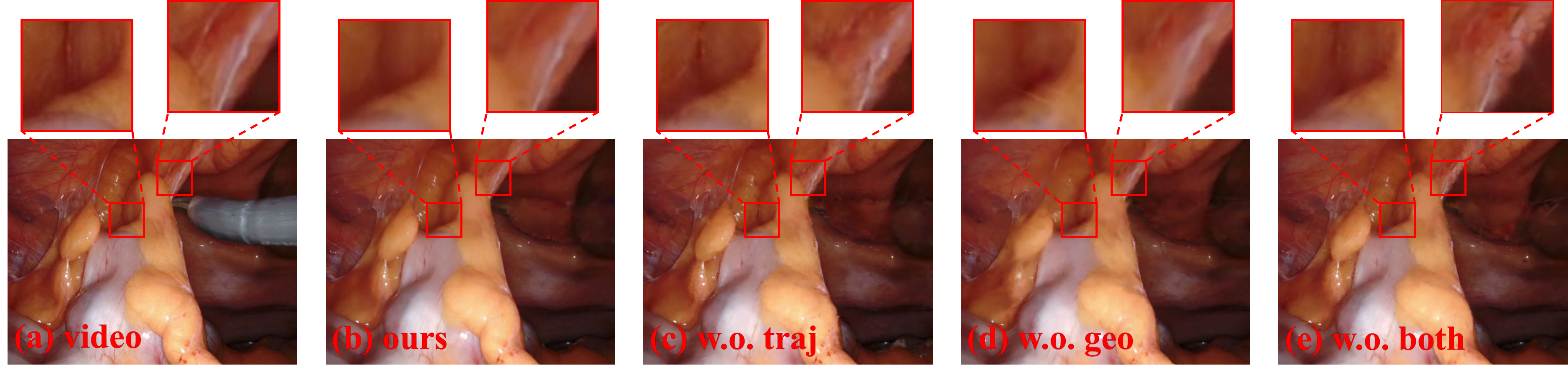}
\end{center}
\caption{Ablation on the trajectory and geometric regulation. (a) is the input video as a reference. (b) is the result of multi-stage optimization. (c), (d) and (e) are the results of optimization without trajectory regularization, geometric regularization, and both, respectively.}
\label{fig:reconexp}
\end{figure*}

Besides, we show Y-T slices across all simulation steps in one case in Fig. \ref{fig:yt}. The epithelial tissue is lifted and then released. Results show that our simulated tissue rapidly reverts to a quiescent state following a quick rebound. Conversely, results from baseline show that the tissue continues to exhibit substantial oscillations post-rebound. This discrepancy is attributed to the proposed Visco-Elastic model which effectively damps out oscillations by simulating the inherent viscoelastic properties of the tissue.

We further provide quantitative comparisons in Tab. \ref{tab:sim}. We reproduce the operation in the input videos and compare the render results with the ground truth. Given the same manipulation, our method shows the best reproduction quality for our physics model behaves more like real tissue, driven by external forces.
\begin{table}[ht]
    \centering
    \footnotesize
    \begin{tabular}{lcccc}
    \hline
    Metric & Ours & Baseline & Pcd & Mesh \\
    \hline
    PSNR$\uparrow$ & \textbf{22.455} & \underline{22.367} & 20.446 & 21.102 \\
    SSIM$\uparrow$ & \textbf{0.7315} & \underline{0.7235} & 0.7068 & 0.6869 \\
    LPIPS$\downarrow$ & \textbf{0.1735} & \underline{0.1759} & 0.2295 & 0.2819 \\
    \hline
    \end{tabular}
    \caption{Quantitative results in terms of simulation quality.}
    \label{tab:sim}
\end{table}

The user study results are shown in Tab. \ref{tab:us}. \name~receives a significant preference across both groups, with a $77.8\%$ preference rate among ordinary viewers and $57.8\%$ among surgeons, significantly outperforming baseline methods. Interestingly, surgeons show a more balanced evaluation, suggesting their professionalism for technical nuances, yet still strongly favored~\name. 

Additionally, we provide the render quality of our canonical model compared with SOTA dynamic tissue reconstruction methods \cite{endonerf,endonerf,lerplane,endogaussian} on EndoNeRF in the supplementary material. Note that our canonical model is for simulation environment preparation instead of reconstruction.

\begin{table}[ht]
\centering
\footnotesize
\begin{tabular}{c|cccc|cc}
\hline
Group & Ours & Baseline & Pcd & Mesh & Ours & w.o. Guide \\
\hline
S & \textbf{57.8\%} & 25.6\% & 10.0\% & 6.7\% & \textbf{65.6\%} & 34.4\% \\
O & \textbf{77.8\%} & 21.3\% & 0.5\% & 0.4\% & \textbf{82.9\%} & 17.1\% \\
\hline
\end{tabular}
\caption{\textbf{User Study}. S stands for surgeon and O for ordinary. The values are the mean percentages of each choice.}
\label{tab:us}
\end{table}

\subsection{Ablation and Analysis}

\paragraph{Multi-Stage Optimization.} In the multi-stage optimization, we employed trajectory regularization and geometric regularization to enhance the geometric consistency. To assess their individual contributions, we conduct ablation studies, and results are shown in Fig. \ref{fig:reconexp}. 
The result without trajectory regularization (c) achieves overall sound quality, but the surfaces and edges of tissues are rough due to the lack of geometric consistency. The result without geometric regularization (d) contains Gaussian kernels with severe anisotropy, which results in burrs on the surface. 
These spiked kernels would result in unrealistic protrusions during surgical simulations.

We also provide a quantitative comparison in Tab.~\ref{tab:recon}. Since the canonical model corresponds to the initial timestamp, rendering metrics are calculated over the first 25 frames. Without geometric regularization, the quality of results is hindered by burrs on the surface. Omitting trajectory regularization doesn't affect the metrics much. As previously discussed, this regularization primarily influences the geometry of the canonical model rather than render quality.

\begin{table}[t]
\centering
\footnotesize
\begin{tabular}{lcccc}
\hline
Metric & Ours & w.o. traj & w.o. geo & w.o. both \\
\hline
% PSNR$\uparrow$ & \textbf{37.114} & \underline{37.036} & 36.582 & 36.750 \\
% SSIM$\uparrow$ & \textbf{0.9772} & \underline{0.9770} & 0.9730 & 0.9754 \\
% LPIPS$\downarrow$ & \textbf{0.0436} & \underline{0.0442} & 0.0505 & 0.0465 \\
PSNR$\uparrow$ & \textbf{37.114} $\pm$ 3.7019 & \underline{37.036} $\pm$ 4.3512 & 36.582 $\pm$ 3.6153 & 36.750 $\pm$ 3.6725 \\
SSIM$\uparrow$ & \textbf{0.9772} $\pm$ 0.0147 & \underline{0.9770} $\pm$ 0.0191 & 0.9730 $\pm$ 0.0143 & 0.9754 $\pm$ 0.0155 \\
LPIPS$\downarrow$ & \textbf{0.0436} $\pm$ 0.0245 & \underline{0.0442} $\pm$ 0.0359 & 0.0505 $\pm$ 0.0230 & 0.0465 $\pm$ 0.0268 \\
\hline
\end{tabular}
\caption{Quantitative results of quality of simulation environment.}
\label{tab:recon}
\end{table}

\paragraph{Physical Parameter Estimation.}
Fig. \ref{fig:yt} shows the results of SurgiSim without video-guided parameter estimation. While the tissue can rebound quickly because of good elastic parameters in the result of SurgiSim, the tissue result without estimation would rebound very slowly with severe damping due to faulty viscous parameters. We also report the participants' preferences on the results with and without physical parameter estimation in Table \ref{tab:us}. 
In both groups, participants show an obvious preference for the results with the inferred physical parameters.

\section{Conclusion}
In this paper, we present \name, an automated and flexible method for transforming monocular surgical videos into simulation-ready scenes, and performing realistic physics surgical simulations within these environments. \name~employs multi-stage optimization with trajectory and anisotropic regularization to construct a geometrically consistent simulation environment. By incorporating a Visco-Elastic deformation model and precise physical parameters estimated from real videos, \name~shows highly realistic tissue deformations during simulation. We hope that \name~could contribute to the development of more diverse and realistic surgical simulations.

\bibliography{loping_temp}
\bibliographystyle{loping_temp}

\newpage
\appendix
\section{Overview}
The contents of this appendix include:
\begin{enumerate}
    \item Attachment Descriptions (Sec.~\ref{App: Supplementary Descriptions}).
    \item Details of Material Point Method (Sec.~\ref{App: Details of Material Point Method}).
    \item More Experimental Results (Sec.~\ref{App: More Experimental Results}).
    \item More Implementation Details (Sec.~\ref{App: More Implementation Details}).
\end{enumerate}

\section{Attachment Descriptions} \label{App: Supplementary Descriptions}
We strongly suggest reviewing our attached \textbf{HTML page}, which contains the following materials:

\begin{enumerate}
    \item \textbf{More Visual Results:} The video page includes demo videos showing different surgical operations across various real surgical scenes. We also provide the complete videos referenced in Fig.~3 of the paper, alongside quantitative comparisons between four methods (Pcd, Mesh, Baseline, SurgiSim) and ablation results without Video Guide.
    
    \item \textbf{User Study Page:} The original interface used in our user study, containing 9 sets of videos for qualitative comparison and Video Guide ablation analysis.
\end{enumerate}

\noindent\textbf{Please click on `\href{https://namaenashibot.github.io/000_SurgiSim/supp/}{index.html}' and select `Videos' to view all attachments.}
% \noindent\textbf{Please click on `\href{./index.html}{index.html}' and select `Videos' to view all attachments.}

\begin{figure}[h!]
\begin{center}
\includegraphics[width=0.475\textwidth]{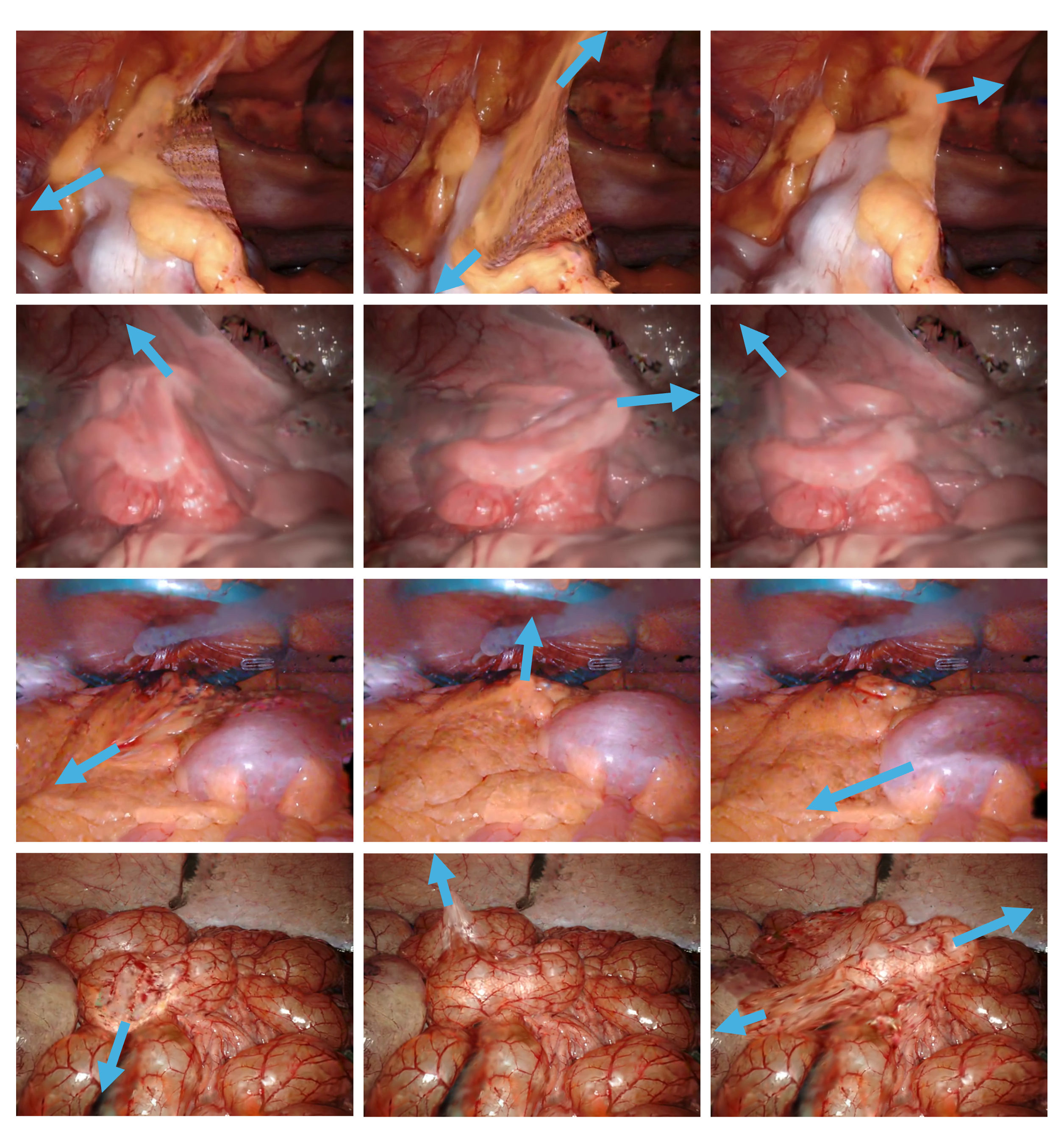}
\end{center}
\caption{Visualization of more simulation cases on different scenes using SurgiSim.}
\label{fig:demo}
\end{figure}

\section{Details of Material Point Method} \label{App: Details of Material Point Method}
The full MPM methods include particle-to-grid (P2G) and grid-to-particle (G2P) to transfer properties between these particles and an Eulerian grid. Following~\citet{mpmsnow,physgaussian}, we use $C^1$ continuous B-spline
kernels for two-way transfer.
The mass and momentum are transferred from particles to grid nodes:
\begin{equation}
m_i = \sum_p m_p \, w_{ip},
\label{eq:p2g_mass}
\end{equation}
\begin{equation}
m_i \mathbf{v}_i = \sum_p m_p \left( \mathbf{v}_p + \mathbf{C}_p (\mathbf{x}_i - \mathbf{x}_p) \right) w_{ip},
\label{eq:p2g_momentum}
\end{equation}
where $m_i$ is the mass at grid node $i$, $m_p\mathbf{v}_p$ are the mass and velocity of particle $p$ and $\mathbf{v}_i$ is the velocity at grid node $i$. $\mathbf{x}_i$ and $\mathbf{x}_p$ are the positions of grid node $i$ and particle $p$, respectively, $w_{ip}$ is the B-spline weighting function between particle $p$ and grid node $i$ and $\mathbf{C}_p$ is the affine velocity matrix of particle $p$, capturing local velocity gradients.

After grid velocities are updated, particle velocities and affine matrices are interpolated from the grid:
\begin{equation}
\mathbf{v}_p^{n+1} = \sum_i w_{ip} \mathbf{v}_i^{n+1},
\label{eq:g2p_velocity}
\end{equation}
\begin{equation}
\mathbf{C}_p^{n+1} = \frac{12}{\Delta x^2(b+1)} \sum_i w_{ip} \mathbf{v}_i^{n+1} (\mathbf{x}_i - \mathbf{x}_p)^\top,
\label{eq:g2p_affine}
\end{equation}
where $\mathbf{v}_p^{n+1}$ is the updated velocity of particle $p$, $\mathbf{v}_i^{n+1}$ is the updated velocity at grid node $i$, $\mathbf{C}_p^{n+1}$ is the updated affine velocity matrix for particle $p$,  $\Delta x$ is the grid spacing and The term $(\mathbf{x}_i - \mathbf{x}_p)^T$ represents the transpose of the position difference vector.

\section{More Experimental Results} \label{App: More Experimental Results}
\subsection{Visualization of Surgical Simulation}
Fig. \ref{fig:demo} presents additional simulation cases, performing different operations on each of the different surgical scenes. The results demonstrate SurgiSim's advanced capability to adapt to diverse new scenes and perform various kinds of operations, including severe pulling and cutting. See the videos for Fig. \ref{fig:demo} and more demos in \href{./index.html}{index.html}.

\subsection{Reconstruction Quality Analysis}
We evaluate SurgiSim's reconstruction capabilities against state-of-the-art dynamic tissue reconstruction methods on the EndoNeRF dataset~\cite{endonerf}. For a fair comparison following the protocol in~\cite{endoreview}, we utilize the original dataset masks and stereo depth information rather than our SAM-refined masks and estimated monocular depth.

Tab.~\ref{tab:metric} presents quantitative results comparing SurgiSim with EndoNeRF~\cite{endonerf}, EndoSurf~\cite{endosurf}, LerPlane~\cite{lerplane}, 4D-GS~\cite{4dgs1}, and EndoGaussian~\cite{endogaussian}. While these metrics primarily evaluate novel-timestamp synthesis capabilities—which is not the primary objective of SurgiSim—our method still achieves compelling results, consistently ranking second best across all metrics. This strong reconstruction performance, achieved as a byproduct of our focus on creating realistic surgical simulation environments, demonstrates SurgiSim's capability to accurately model tissue deformations across video frames, proving its powerful ability to extract high-quality canonical scenes from dynamic inputs.

\begin{table*}[t!]
\centering
\small
\begin{tabular}{lcccccc}
\toprule
Metric & SurgiSim & EndoNeRF & EndoSurf & LerPlane & 4D-GS & EndoGaussian \\
\midrule
PSNR$\uparrow$ & 35.490 & 27.077 & 34.795 & 34.643 & 22.832 & 36.31 \\
SSIM$\uparrow$ & 0.966 & 0.900 & 0.945 & 0.922 & 0.827 & 0.971 \\
LPIPS$\downarrow$ & 0.056 & 0.107 & 0.119 & 0.072 & 0.368 & 0.050 \\
\bottomrule
\end{tabular}
\caption{Quantitative comparison of reconstruction quality on the EndoNeRF dataset. While our method is not specifically designed for novel-timestamp synthesis, it achieves competitive performance across all metrics.}
\label{tab:metric}
\end{table*}

\section{More Implementation Details} \label{App: More Implementation Details}
\subsection{Training and Performance}
The training is per scene. For each input video, it takes less than 4 minutes to build a canonical scene from the input video with Surface Thickening (Sec. 3.2) done. For physical parameter estimation, because we train in a rolling manner (Sec. 3.3), the time complexity concerning the length of the operation is $O(n^2)$. The mean optimization time is 10 minutes, but the time can reach 20 minutes for long sequences. When performing novel simulations after optimization, the simulation speed can reach 7 fps by setting the simulation step duration 10 times longer for fewer steps.

All experiments were conducted on a machine equipped with a Core i7-13700K CPU and a single NVIDIA RTX 4090 GPU, running Ubuntu 24.04. The code will be made public to promote the virtual surgery. 

\subsection{Details on User Study}
To evaluate the fidelity of our simulations, we conducted a user study involving 68 participants including both surgeons and laypersons. These participants were categorized into two distinct groups: the Surgeons group, consisting of 44 board-certificated surgeons, and the Ordinary group, comprising 24 laypersons with no medical background.

The study was structured into two parts. In the first part, participants were tasked with selecting one most realistic simulation from four options, results from four different methods. In the second part, they were required to choose one superior simulation between the two presented results. Each participant was exposed to nine sets of simulation results, with the order of the sets randomized to prevent order bias. Before analysis, we executed a basic data-cleaning process to remove any invalid or outlier responses.

We provide the page used for our user study in the supplement materials. To view the page, just click on the `\href{https://namaenashibot.github.io/000_SurgiSim/supp/}{index.html}' and select `User Study'.
% We provide the page used for our user study in the supplement materials. To view the page, just click on the `\href{./index.html}{index.html}' and select `User Study'.

\subsection{Future Work}
In the simulation environment setup, our method struggled to handle the invisible portions on the sides of the tissues, leading to imperfections in the texture generated by our thickening approach. Similarly, the texture on the exposed areas after cutting still lacked sufficient realism and required manual correction. In the future, we hope to use diffusion or the large reconstruction model to fix these textures.
Our method does not yet support more complicated operations like topological inversions of structure. In the future, we hope to build a geometry-aware MPM method based on 3D scene understanding techniques.

\end{document}